\documentclass[10pt,twocolumn]{IEEEtran} 

\usepackage{fullpage,graphicx,psfrag,url,setspace}
\usepackage{cite}
\usepackage{epsf}
\usepackage{color}
\usepackage{slashbox}
\usepackage[small,bf]{caption}
\usepackage{amsmath,verbatim,amsthm,amssymb,amsfonts,amscd, graphicx,algorithmic}

\usepackage{lipsum}

\usepackage{stmaryrd}
\usepackage{array}
\usepackage{placeins}
\usepackage{algorithm, algorithmic}

\usepackage{textcomp}

\newcommand{\ben}{\begin{equation}}
\newcommand{\bs}{\begin{split}}
\newcommand{\es}{\end{split}}

\newcommand{\een}{\end{equation}}
\newtheorem{thm}{Theorem}

\title{\LARGE{Sequential Changepoint Approach for Online Community Detection}}

\author{David Marangoni-Simonsen \thanks{David Marangoni-Simonsen (Email: dmarangonisi@gatech.edu), and Yao Xie (Email: yao.xie@isye.gatech.edu) are with the H. Milton Stewart School of Industrial and Systems Engineering, Georgia Institute of Technology, Atlanta, GA.} and Yao Xie 
}

\begin{document}
\maketitle

\begin{abstract}
We present new algorithms for detecting the emergence of a community in large networks from sequential observations. The networks are modeled using Erd\H{o}s-Renyi random graphs with edges forming between nodes in the community with higher probability.  Based on statistical changepoint detection methodology, we develop three algorithms: the Exhaustive Search (ES), the mixture, and the Hierarchical Mixture (H-Mix) methods. Performance of these methods is evaluated by the average run length (ARL), which captures the frequency of false alarms, and the detection delay. Numerical comparisons show that the ES method performs the best; however, it is exponentially complex. The mixture method is polynomially complex by exploiting the fact that the size of the community is typically small in a large network. However, it may react to a group of active edges that do not form a community. This issue is resolved by the H-Mix method, which is based on a dendrogram decomposition of the network. 
We present an asymptotic analytical expression for ARL of the mixture method  when the threshold is large. Numerical simulation verifies that our approximation is accurate even in the non-asymptotic regime. Hence, it can be used to determine a desired threshold efficiently. Finally, numerical examples show that the mixture and the H-Mix methods can both detect a community quickly with a lower complexity than the ES method.
\end{abstract}


\section{Introduction}

Community detection within a network arises from a wide variety of applications, including advertisement generation to cancer detection \cite{fortunato13 , newman04,pandit2011,kawadia2012,chen2013}.
%
These problems often consist of some graph $\mathcal{G}$ which contains a community $\mathcal{C} \subset \mathcal{G}$ where $\mathcal{C}$ and $\mathcal{G} \backslash \mathcal{C}$ differ in some fundamental characteristic, such as the frequency of interaction (see \cite{radicchi04} for more details). Often this community $\mathcal{C}$ is assumed to be a clique, which is a network of nodes in which every two nodes are connected by an edge. 
Here we consider the {\it statistical community detection} problem, where the observed edges are noisy realizations of the true graph structures, i.e., the observations are random graphs.

Community detection problems can be divided into either one-shot \cite{leung09, leskovec10, bhattacharyya13, verzelen13, cai14} or dynamic categories \cite{koujaku13, online_comm_Zhejiang2013, duan13}. The more commonly considered one-shot setting assumes observations from static networks. The dynamic setting is concerned with sequential observations from possibly dynamic networks, and has become increasingly important since such scenarios become prevalent in social networks \cite{online_comm_Zhejiang2013}. These dynamic categories can be further divided into networks with (1) structures that either continuously change over time \cite{koujaku13}, or (2) structures that change abruptly after some changepoint $\kappa$ \cite{duan13}, the latter of which will be the focus of this paper. 

In online community detection problems, due to the real time processing requirement, we cannot simply adopt the exponentially complex algorithms, especially for large networks.
Existing approaches for community detection can also be categorized into parametric \cite{kolar04, lambiotte10,barbieri12,sharpnack12} and non-parametric methods \cite{wasserman08}. 
However, many such methods \cite{wasserman08, sharpnack12} rely on the data being previously collected and would not be appropriate for streaming data.

Existing online community detections algorithms are usually based on heuristics (e.g. \cite{leung09}). It is also recognized in \cite{arias13} that there has been tenuous theoretical research regarding the detection of communities in static networks. The community detection problem in \cite{arias13} was therefore cast into a hypothesis testing framework, where the null hypothesis is the nonexistence of a community, and the alternative is the presence of a {\it single} community. They model networks using an Erd\H{o}s-Renyi graph structure due to its comparability to a scale-free network. Based on this model, they derive scan statistics which rely on counting the number of edges inside a subgraph \cite{verzelen13}, and establish the fundamental detectability regions for such problems. 

In this paper, we propose a sequential changepoint detection framework for detecting an abrupt emergence of a {\it single} community using sequential observations of random graphs. We also adopt the Erd\H{o}s-Renyi model, but our methods differ from \cite{verzelen13} in that we use a sequential hypothesis testing formulation and the methods are based on  sequential likelihood ratios, which have statistically optimal properties. From the likelihood formulations, three sequential procedures are derived: the Exhaustive Search (ES), the mixture, and the Hierarchical Mixture (H-Mix) methods. The ES method performs the best but it is exponentially complex even if the community size is known; the mixture method is polynomially complex and exploits the fact that the size of the community inside a network is typically small. A limit of the mixture method is that it raises a false alarm due to a set of highly active edges that do not form a community. The H-Mix method addresses this problem by imposing a dendrogram decomposition of the graph. The performance of the changepoint detection procedures are evaluated using the average-run-length (ARL) and the detection delay. We derived a theoretical asymptotic approximation of the ARL of the mixture method, which was numerically verified to be accurate even in the non-asymptotic regime. Hence, the theoretical approximation can be used to determine the detection threshold efficiently. The complexity and performance of the three methods are also analyzed using numerical examples. 

This paper is structured as follows. Section \ref{sec:formulation} contains the problem formulation. Section \ref{sec:sequential} presents our methods for sequential community detection. Section \ref{sec:theoretical} explains the theoretical analysis of the ARL of the mixture model. Section \ref{sec:num_eg} contains numerical examples for comparing performance of various methods, and Section \ref{sec:con} concludes the paper. 

\section{Formulation}\label{sec:formulation}

\begin{figure}[h!]
  \centering
\includegraphics[width=0.65\linewidth]{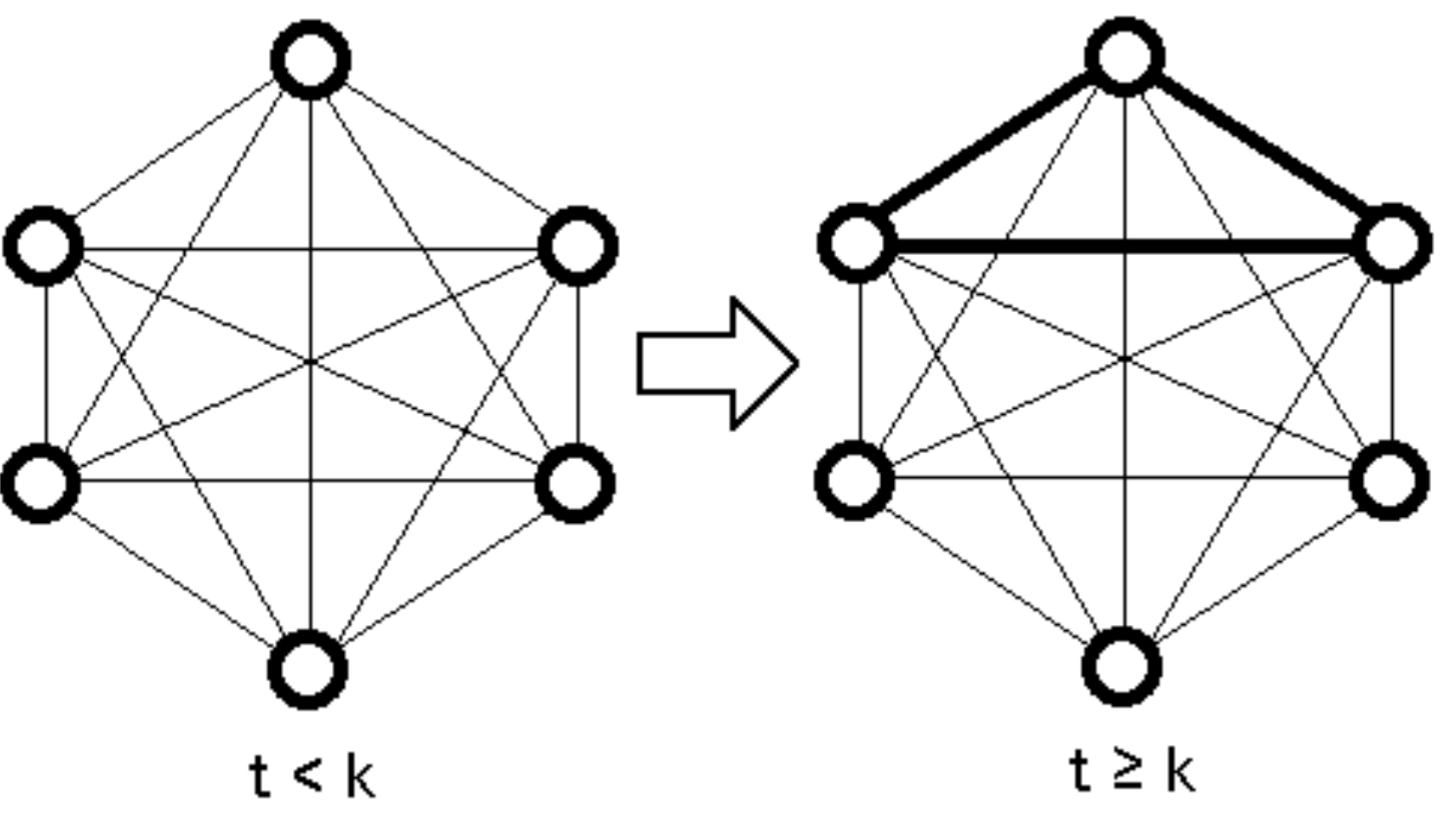} 
  \caption{The graph structure displays interactions between edges in a 6 node network. Assume that when there is no community present, edges between nodes form with probability $p_0$ (denoted by light lines). Starting from time $\kappa$, a community forms and the nodes in the community interact with each other with a higher probability $p_1$ (denoted by bold lines).}\label{changepoint_basis_3}
\end{figure}

\begin{figure}[h!]
  \centering
\includegraphics[width=0.95\linewidth]{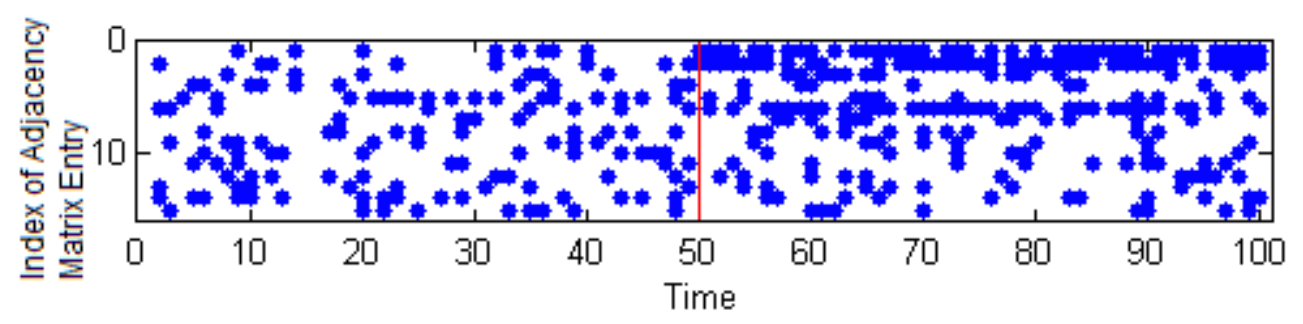} 
  \caption{An instance of adjacency matrices observed over time for the 6 node network in Figure \ref{changepoint_basis_3}. Each column in the figure represents the lower triangle entries of the adjacency matrix (the matrix is symmetric since we assume the graph is undirectional).  
At time $\kappa=50$ (indicated by the red line) a community forms between nodes 1, 2, and 3 and the edges between these three nodes are formed with a higher probability. }
  \label{fig:spy_plot_3}
\end{figure}

Assume a network with $N$ nodes and an observed sequence of adjacency matrices over time $X_1, X_2, \ldots$ with $X_t \in \mathbb{R}^{N\times N}$, where $X_i$ represents the interaction of  these nodes at time $i$. Also assume when there is no community, there are only random interactions between all nodes in the network with relatively low frequency. 
There may exist an (unknown) time at which a community emerges and nodes inside the community have much higher frequencies of interaction. Figures~\ref{changepoint_basis_3} 
and \ref{fig:spy_plot_3} illustrate such a scenario.

We formulate this problem as a sequential changepoint detection problem. The null hypothesis is that the graph corresponding to the network at each time step is a realization of an Erd\H{o}s-Renyi random graph, i.e., edges are independent Bernoulli random variables that take values of 1 with probability $p_0$ and values of 0 with probability $1-p_0$. Let $[X]_{ij}$ denote the $ij$th element of a matrix $X$, then
\begin{eqnarray}
[X_t]_{ij} =\left\{ \begin{array}{ll}
1 & \mbox{w. p. } p_0\\
0 & \mbox{otherwise}
\end{array}\right. \quad \forall (i, j).
\label{eq:hyp_base}
\end{eqnarray}
The alternative hypothesis is that there exists an unknown time $\kappa$ such that after $\kappa$, an {\it unknown} subset of nodes $\mathcal{S}^*$ in the graph form edges between community nodes with a higher probability $p_1$, $p_1 > p_0$: 
\begin{eqnarray}
[X_t]_{ij} =\left\{ \begin{array}{ll}
1 & \mbox{w. p. } p_1\\
0 & \mbox{otherwise}
\end{array}\right. \quad \forall i, j \in \mathcal{S^*}, \quad t > \kappa
\label{eq:hyp_in}
\end{eqnarray}
and for all other connections
\begin{eqnarray}
[X_t]_{ij} =\left\{ \begin{array}{ll}
1 & \mbox{w. p. } p_0\\
0 & \mbox{otherwise}
\end{array}\right. \forall i \notin \mathcal{S}^* \mbox{ or } j \notin \mathcal{S}^*, \quad t > \kappa.
\label{eq:hyp_out}
\end{eqnarray}
We assume that $p_0$ is known, as it is a baseline parameter which can be estimated from historic data. We will consider both cases when $p_1$ is either unknown or known. Our goal is to define a stopping rule $T$ such that for a large {\it average-run-length (ARL)} value, $\mathbb{E}^\infty\{T\}$, the {\it expected detection delay} $\mathbb{E}^\kappa\{T-\kappa|T>\kappa\}$ is minimized. Here $\mathbb{E}^\infty$ and $\mathbb{E}^\kappa$ consecutively denote the expectation when there is no changepoint, and when the changepoint occurs at time $\kappa$. 

\section{Sequential community detection} \label{sec:sequential}

Define the following statistics for edge $(i, j)$ and assumed changepoint time $\kappa=k$ for observations up to some time $t$,
\begin{equation}
 U_{k, t}^{(i,j)} =  \sum_{m=k+1}^t [X_m]_{ij} \log \left(\frac{p_1}{p_0}\right) +  
  (1-[X_m]_{ij})\log \left(\frac{1-p_1}{1-p_0}\right),
 \label{U_def}
\end{equation}
Then for a given changepoint time $\kappa = k$ and a community $\mathcal{S}$, we can write the log-likelihood ratio for (\ref{eq:hyp_base}), (\ref{eq:hyp_in}) and (\ref{eq:hyp_out}) as follows:
\begin{equation}
\begin{split}
 \mathcal{\ell} (\kappa = k | \mathcal{S})  & \triangleq \log \left( \prod_{m = k+1}^t \prod_{(i,j) \in \mathcal{S} }\frac{p_1^{[X_m]_{ij}} (1-p_1)^{1-[X_m]_{ij}}}{p_0^{[X_m]_{ij}} (1-p_0)^{1-[X_m]_{ij}}} \right) \\
& = \sum_{(i, j) \in \mathcal{S}} U_{k, t}^{(i,j)}.
 \end{split}
\label{eq:log_like_eq}
\end{equation}

Often, the probability $p_1$ of two community members interacting is unknown since it typically represents an anomaly (or new information) in the network. In this case, $p_1$ can be replaced by its maximum likelihood estimate, which can be found by taking the derivative of $\ell (\kappa = k | \mathcal{S})$ (\ref{eq:log_like_eq}) with respect to $p_1$, setting it equal to $0$ and solving for $p_1$:
\begin{equation}
\begin{split}
\widehat{p}_1 & = \frac{2 }{| \mathcal{S} |  (| \mathcal{S} |-1)(t-k) } \sum_{(i, j) \in \mathcal{S}} \sum_{m=k+1}^t [X_m]_{ij}.
\label{eq:p1_hat_equation}
\end{split}
\end{equation}
where $|\mathcal{S}|$ is the cardinality of a set $\mathcal{S}$. In the following procedures, whenever $p_1$ is unknown, we replace it with $\widehat{p}_1$.

\subsection{Exhaustive Search (ES) method}

First consider a simple sequential community detection procedure assuming  the size of the community, $|\mathcal{S}^*|=s$, and $p_1$ are known. 
The test statistic is the maximum log likelihood ratio (\ref{eq:log_like_eq}) over all possible sets $\mathcal{S}$ and all possible changepoint locations in a time window $k \in [t-m_1, t-m_0]$. Here $m_0$ is the start and $m_1$ is the end of the window. This window limits the complexity of the statistic, which would grow linearly with time if a window is not used. The stopping rule is to claim there has been a community formed whenever the likelihood ratio exceeds a threshold $b > 0$ at certain time $t$.
Let $\llbracket N \rrbracket \triangleq \{ 1, \dots , N\}$. The exhaustive search (ES) procedure is given by the following
\begin{equation}
T_{\rm ES} = \inf \{ t: \max_{t-m_1\leq k\leq t-m_0}  \max_{\mathcal{S}\subset \llbracket N \rrbracket:  |\mathcal{S}| = s}    \sum_{(i, j) \in \mathcal{S}} U_{k, t}^{(i,j)} \geq b \}, 
\label{T_ES_def}
\end{equation}
where $b > 0$ is the threshold. 

Note that the testing statistic in (\ref{T_ES_def}) searches over all  $2^s$ possible communities, which is exponentially complex in the size of the community $s$. One fact that alleviates this problem is when $p_1$ is known, there exists a recursive way to calculate the test statistic $\max_{k\leq t}    \sum_{(i, j) \in \mathcal{S}} U_{k, t}^{(i,j)}$ in (\ref{T_ES_def}), called the CUSUM statistic \cite{Siegmund1985}.  
%
For each possible $\mathcal{S}$, when $m_0 = 0$, we calculate
\begin{equation}
 W_{\mathcal{S}, t+1} =\max\{ W_{\mathcal{S},t} + \sum_{(i,j) \in \mathcal{S}} U_{t,t+1}^{(i,j)}, 0 \},
\label{eq:cusum_equation} 
\end{equation}
with
$
W_{\mathcal{S},t} = 0,
$
and the detection procedure (\ref{T_ES_def}) is equivalent to:
\begin{equation}
\begin{split}
& T_{\rm ES} = \inf \{ t: \max_{\mathcal{S}\subset \llbracket N \rrbracket:  |\mathcal{S}| = s}  W_{\mathcal{S}, k} \geq b \},
\end{split}
\end{equation}
where $b$ is the threshold. When $p_1$ is unknown, however, there is no recursive formula for calculating the statistic, due to a nonlinearity resulting from substituting $\widehat{p}_1$ for $p_1$.

\subsection{Mixture method}
\label{sec:mixture_model}
The mixture method avoids the exponential complexity of the ES method by introducing a simple probabilistic mixture model, which exploits the fact that typically the size of the community is small, i.e. $|\mathcal{S}^*| /N \ll 1$. It is motivated by the mixture method developed for detecting a changepoint using multiple sensors \cite{xie13} and detecting aligned changepoints in multiple DNA sequences \cite{SiegmundYakirZhang2011}. 
The mixture method does not require knowledge of the size of the community $|\mathcal{S}^*|$. 


We assume that each edge will happen to be a connection between two nodes inside the community with probability $\alpha$, and use i.i.d. Bernoulli $Q_{ij}$ indicator variables that take on a value of 1 when node $i$ and node $j$ are both inside the community and $0$ otherwise:  
\begin{eqnarray}
Q_{ij} =\left\{ \begin{array}{ll}
1 & \mbox{w. p. } \alpha\\
0 & \mbox{otherwise}
\end{array}\right. \quad \forall i, j \in \mathcal{S}^*.
\end{eqnarray}
Here $\alpha$ is a guess for the fraction of edges that belong to the community. Let
\begin{equation}
\begin{split}
& h(x) \triangleq \log \{ 1 - \alpha + \alpha \exp (  x )\}.
\label{eq:h_equation}
\end{split}
\end{equation}
With such a model, the likelihood ratio can be written as
\begin{equation}
\begin{split}
& \ell ( \kappa = k | \mathcal{S})  \\
= & \log \prod_{1 \leq i < j \leq N} \mathbb{E}_{Q_{ij}} [  (1-Q_{ij})  + \\
&Q_{ij}  \prod_{m = k+1 }^t  \frac{p_1^{[X_m]_{ij}} (1- p_1) ^{1- [X_m]_{ij}} }{p_0^{[X_m]_{ij}} (1- p_0) ^{1- [X_m]_{ij}} } ]  \\
= & \sum_{1\leq i < j \leq N} h(U_{k, t}^{(i, j)}),
\end{split}
\end{equation}
and the mixture method detects the community using a stopping rule:
\begin{equation}
\begin{split}
& T_{\rm Mix} = \inf\{t: \max_{t-m_1\leq k\leq t-m_0} \sum_{1\leq i < j \leq N} h(  U_{k, t}^{(i,j)} ) \geq b\} ,
\end{split}
\label{T_mix}
\end{equation}
where $b>0$ is the threshold. Here $h(x)$ can be viewed as a soft-thresholding function \cite{xie13} that selects the edges which are more likely to belong to the community. 

However, one problem with a mixture method is that its statistic can be gathered from edges that do not form a community. Figure \ref{fig:mixture_model_failure_4} below displays two scenarios where the mixture statistics will be identical, but Figure \ref{fig:mixture_model_failure_4}(b) does not correspond to a network forming a community. To solve this problem, next we introduce the hierarchical mixture method. 

\begin{figure}[h!]
\begin{center}
\begin{tabular}{cc}
      \includegraphics[width=0.3\linewidth]{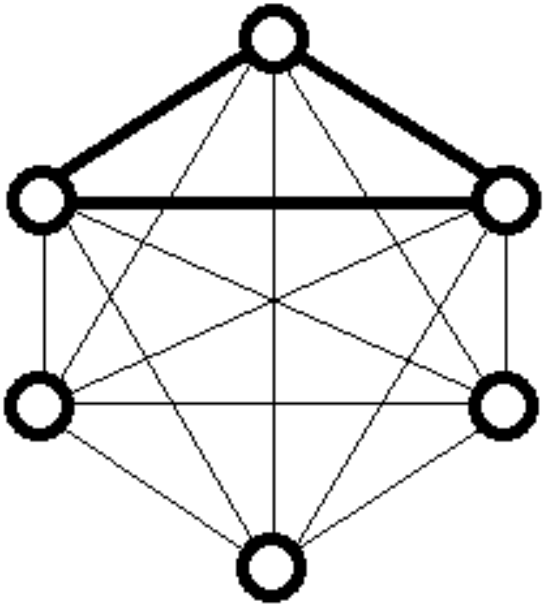} & \includegraphics[width=0.3\linewidth]{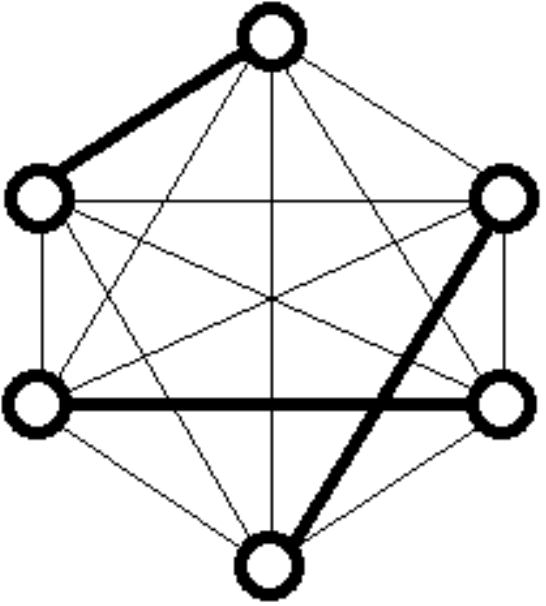} \\
(a) & (b)
\end{tabular}
\end{center}
  \caption{(a): a community where all nodes in the community are connected with a higher probability than under the null hypothesis. (b): a model which would output the same mixture statistic that does not correspond to a community.}
  \label{fig:mixture_model_failure_4}
\end{figure}

\subsection{Hierarchical Mixture method (H-Mix) }

In this section, we present a hierarchical mixture method (H-Mix) that takes advantage of the low computational complexity of the mixture method in section~\ref{sec:mixture_model} while enforcing the statistics to be calculated only over meaningful communities.
Hence, the H-Mix statistic is robust to the non-community interactions displayed in Figure~\ref{fig:mixture_model_failure_4}. 
The H-Mix method requires the knowledge (or a guess) of the size of the community $|\mathcal{S}^*|$.

\begin{figure}[h!]
  \centering
      \includegraphics[width=0.9\linewidth]{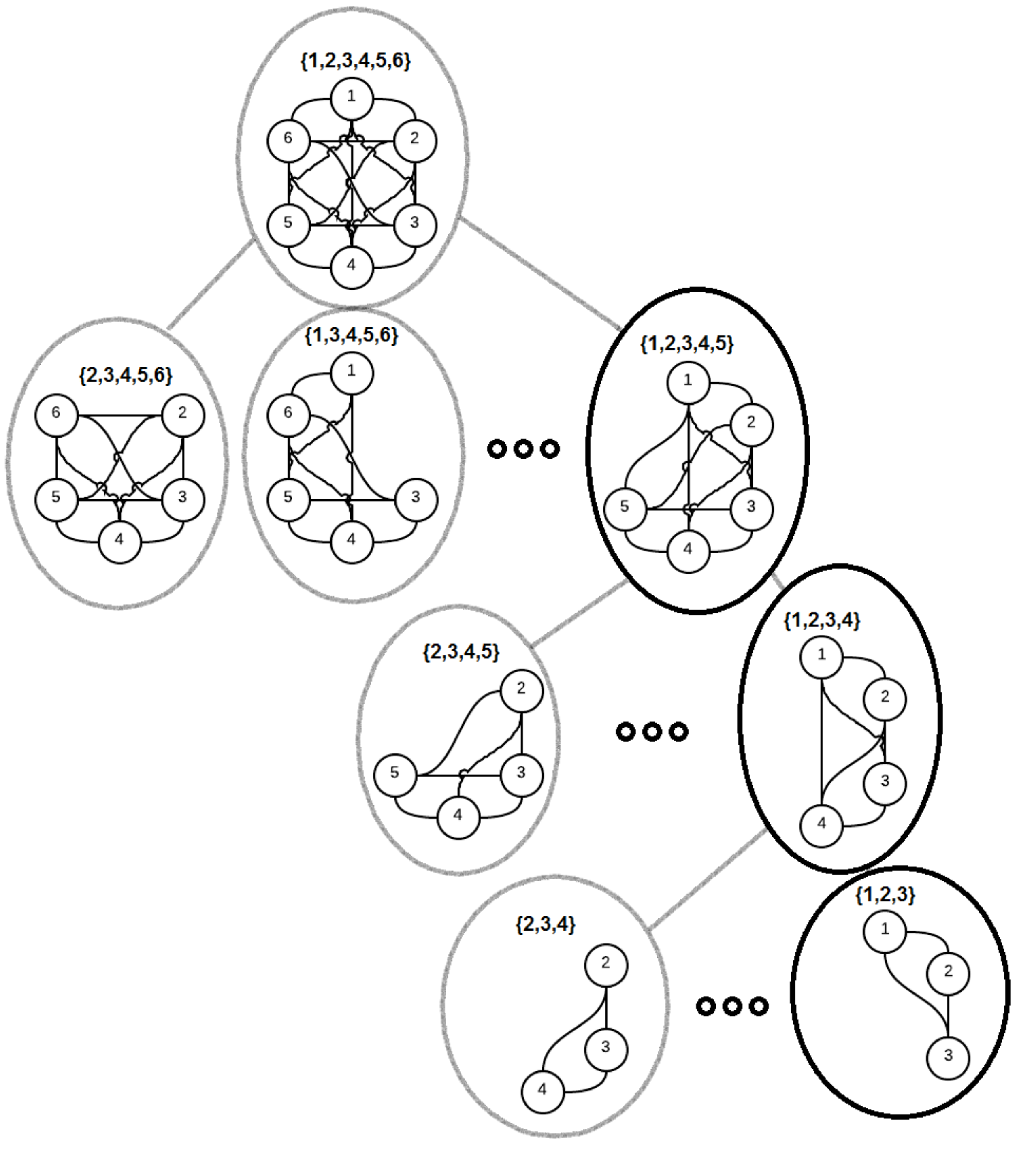}
  \caption{A possible outcome for a single instance with $s=3$ and $N=6$, in which a community is found consisting of nodes 1, 2, and 3. The original set of nodes consisted of the set $\{1, 2, 3, 4, 5, 6\}$, and the H-Mix method followed the dendrogram down the node sets with darker outlines (which had the highest mixture method statistic amongst their group) until at the $s=3$ level the set $\{1, 2, 3\}$ was selected.
  }
  \label{fig:online_community_detection_forced_conn_alg_pic}
\end{figure}

The H-Mix method enforces the community structure by constructing a dendrogram decomposition of the network, which is a hierarchical partitioning of the graphical network \cite{krishnamurthy13}. The hierarchical structure provided by dendrogram enables us to systematically remove nodes from being considered for the community. Suppose a network has a community of size $s$. Starting from the root level with all nodes belonging to the community, each of the nodes in the dendrogram tree decomposition is a subgraph of the entire network that contains all but one node. Then the mixture statistic from (\ref{T_mix}) is evaluated for each subgraph: using $h(x)$ defined in 
(\ref{eq:h_equation}), for a given set of nodes $\mathcal{S}_0$ and a hypothesized changepoint location $k$, the mixture statistic is calculated as
\begin{equation}
M (\mathcal{S}_0) =  \sum_{ (i,j) \in \mathcal{S}_0} h \left( U_{k, t}^{(i,j)}  \right).
\label{mix_stat}
\end{equation}
We iteratively select the subgraph with the highest mixture statistic value, since it indicates that the associated node removed is most likely to be a non-member of the community and will be eliminated from subsequent steps. The algorithm repeats until there are only $s$ nodes remaining in the subgraph. Denote the mixture statistic for the selected subgraph as $P_k$. Then $\{P_k\}_{k=1}^t$ is a series of test statistics at each hypothesized changepoint location $k$. Finally, the H-Mix method is given by
\begin{equation}
\begin{split}
& T_{\rm H-Mix}= \inf\{t: \max_{t-m_1\leq k\leq t-m_0} P_k \geq b\},
\end{split}
\end{equation}
where $b$ is the threshold. The idea for a dendrogram decomposition is similar to the edge removal method \cite{newman04}, and here we further combine it with the mixture statistic. Figure~\ref{fig:online_community_detection_forced_conn_alg_pic} illustrates the procedure described above and Algorithm \ref{alg:hier_alg} summarizes the H-Mix method.

\begin{algorithm}[h!]
\caption{Hierarchical Mixture Method} \label{alg:hier_alg}
\begin{algorithmic}[1]
\STATE Input: $\{X_m\}_{m=1}^t, X_m \in \mathbb{R}^{N \times N}$
\STATE Output: $\{P_k\}_{k=1}^t \in \mathbb{R}^t $, a set of statistics for each hypothesized changepoint location $k$.
\FOR{$k=1 \to t$}
\STATE $\mathcal{S} = \llbracket N \rrbracket$
\WHILE{$|\mathcal{S}|>s$}
\STATE $i^* = \text{argmax}_{i \in \mathcal{S}} M \left( \mathcal{S} \text{\textbackslash} \{i\} \right)$
\STATE $\mathcal{S}  = \mathcal{S} \text{\textbackslash} \{i^*\}$
\ENDWHILE
\STATE $P_k = M ( \mathcal{S})$
\ENDFOR
\end{algorithmic}\label{MMFCalg}
\end{algorithm}


\subsection{Complexity}

\begin{table}[h!]
\begin{center}
\caption{Complexities of algorithms under various conditions.}\label{table:complexity_table}
  \begin{tabular}{ |c|c|c|}
    \hline
     & $s \ll N/2$ & $s \sim N/2$ \\ \hline
    ES &  $\mathcal{O} ( {N^{ s }} )$ & $\mathcal{O} ( {2^{ \frac{s}{2} }} )$\\ \hline
    Mixture  & $\mathcal{O} (N^2)$ & $\mathcal{O} (N^2)$\\ \hline
H-Mix  & $\mathcal{O} (N^4)$ & $\mathcal{O} (N^4)$\\ \hline
  \end{tabular}
\end{center}
\end{table}

In this section, the algorithm complexities will be analyzed and the complexities are summarized in Table \ref{table:complexity_table}. The derivation of these complexities are explained as follows. Given a known subset size $s$ and at a given changepoint location $k$ and current time $t$, evaluating the ES test statistic requires in $\mathcal{O} ( {N \choose s } )$ operations. Using Stirling's Approximation $\log( {N \choose s }) \sim N H(\frac{s}{N})$, where the entropy function is $H(\epsilon) = -\epsilon \log(\epsilon) - (1-\epsilon) \log( 1- \epsilon)$, the complexity of evaluating ES statistic is $\mathcal{O} ((\frac{N}{s})^{s } (\frac{N}{N- s})^{N - s } )$. This implies that for $s \ll N/2$, the complexity will be approximately polynomial $\mathcal{O} ( {N^{s}} )$. However, a worst case scenario occurs when $s \sim N/2$, as the statistic must search over the greatest number of possible networks and the complexity will consequently be $\mathcal{O} ( {2^{ \frac{s}{2} }} )$ which is exponential in $s$. 


The mixture method only uses the sum of all edges and therefore the complexity is $\mathcal{O} (N^2)$. Unlike the exhaustive search algorithm, the mixture model has no dependence on the subset size $s$, by virtue of introducing the mixture model with parameter $\alpha$.


On the $i$th step, the H-Mix algorithm computes $N+1-i$ mixture statistics, and there are $N-s$ steps required to reduce the node set to $s$ final nodes. 
%
Therefore the total complexity is on the order of $\sum_{i=1}^{N-s} \sum_{j=1}^{N+1-i} (N+1-i)^2 $, which is further reduced to $\mathcal{O} (N^4)$ if it is assumed that $s \ll N$.

\section{Theoretical Analysis for mixture method}\label{sec:theoretical}

In this section, we present a theoretical approximation for the ARL of the mixture method when $p_1$ is known using techniques outlined as follows. In \cite{SiegmundYakirZhang2011} a general expression for the tail probability of scan statistics is given, which can be used to derive the ARL of a related changepoint detection procedure. For example, in \cite{xie13}  a generalized form for ARL was found using the general expression in \cite{SiegmundYakirZhang2011} . The basic idea is to relate the probability of stopping a changepoint detection procedure when there is no change, $\mathbb{P}^\infty\{T\leq m\}$, to the tail probability of the maxima of a random field: $\mathbb{P}^\infty\{S \geq b\}$, where $S$ is the statistic used for the changepoint detection procedure, $b$ is the threshold, and $\mathbb{P}^\infty$ denotes the probability measure when there is no change. Hence, if we can write $\mathbb{P}^\infty\{S \geq b\} \approx m \lambda$ for some $\lambda$, by relying on the assumption that the stopping time is asymptotically exponentially distributed when $b\rightarrow \infty$, we can find the ARL is $1/\lambda$. 
%
However, the analysis for the mixture method here differs from that in \cite{xie13} in two major aspects: (1) the detection statistics here rely on a binomial random variable, and we will use a normal random variable to approximate its distribution; (2) the change-of-variable parameter $\theta$ depends on $t-k$ and, hence, the expression for ARL will be more complicated than that in \cite{xie13}. 

\begin{thm}\label{thm1}
When $b\rightarrow \infty$, an upper bound to the ARL  $\mathbb{E}^\infty[T_{\rm mix}]$ of the mixture method with known $p_1$ is given by:
\begin{equation}
\begin{split}
\mbox{ARL}_{\rm UB}  
=  \left[ \int_{\sqrt{2N/m_1}}^{\sqrt{2N/m_0}} \frac{y \nu^2(y\sqrt{\gamma(\theta_y)})}{H(N, \theta_y)} dy \right]^{-1},
\end{split}
\label{eq:ARL_integration}
\end{equation}
and a lower bound to the ARL is given by:
\begin{equation}
\mbox{ARL}_{\rm LB} = \left[\sum_{\tau = m_0}^{m_1} 
\frac{2N  \nu^2(2N\sqrt{\gamma(\theta_\tau)}/\tau^2) }{\tau^2 H(N, \theta_\tau)}
\right]^{-1}, \label{ARL_LB}
\end{equation}
where 
\begin{equation}
\begin{split}
c_0 & = \log (p_1/p_0),  \quad c_1 = \log [(1-p_1)/(1-p_0)],\\
g_{\tau} (x) &=   \tau[p_0  (c_0 - c_1)   + 
 c_1] + 
 x \sqrt{\tau (c_0 -c_1)^2 p_0 (1-p_0)},\\
h_{\tau}(x)  & \triangleq h(g_{\tau} (x)),  \mbox{ for $h(x)$ defined in (\ref{eq:h_equation})}, \\
\psi_\tau(\theta) & = \log \mathbb{E}\{e^{\theta h_{\tau} (Z)  }\}, \\
\dot{\psi}_\tau(\theta) & = \frac{\mathbb{E}\{ h_{\tau} (Z) e^{\theta h_{\tau} (Z) }\} }{\mathbb{E}\{e^{\theta h_{\tau} (Z) }\}}, \\
\ddot{\psi}_\tau(\theta) & = \frac{\mathbb{E}\{ h_{\tau}^2 (Z) e^{\theta h_{\tau} (Z) }\} }{\mathbb{E}\{e^{\theta h_{\tau} (Z) }\}}  - \frac{ (\mathbb{E}\{ h_{\tau} (Z) e^{\theta h_{\tau} (Z) }\} )^2 }{ (\mathbb{E}\{e^{\theta h_{\tau} (Z) }\} )^2 },\\
\gamma(\theta) &= \frac{1}{2} \theta^2  \mathbb{E}\{  [ \dot{h}_{\tau}(Z) ]^2 \exp \{ \theta h_{\tau}(Z)  - \psi_{\tau} (\theta) \}, \\
\theta_\tau & \mbox{ is solution to } \dot{\psi}_{\tau}(\theta_\tau) = b/N,\\
H(N, \theta_\tau) & = \frac{\theta [2\pi  \ddot{\psi}(\theta_\tau)]^{1/2}}{\gamma^2(\theta_\tau) \sqrt{N}}e^{N[\theta\dot{\psi}(\theta_\tau) - \psi(\theta_\tau)]},
 \end{split}
\end{equation}
and $\dot{f}$ and $\ddot{f}$ denote the first and second order derivatives of  a function $f$, $Z$ is a normal random variable with zero mean and unit variance, the expectation is with respect to $Z$, and the special function $\nu(x)$ is given by \cite{xie13} 
\[
\nu(x) \approx \frac{(2/x)[\Phi(x/2) - 1/2]}{(x/2)\Phi(x/2) + \phi(x/2)}.
\]
Here $\theta_\tau$ is the solution to \[\dot{\psi}_{\tau}(\theta_\tau) = b/N.\]

\end{thm}

\begin{figure}[h!]
\begin{center}
\begin{tabular}{cc}
\includegraphics[width=0.45\linewidth]{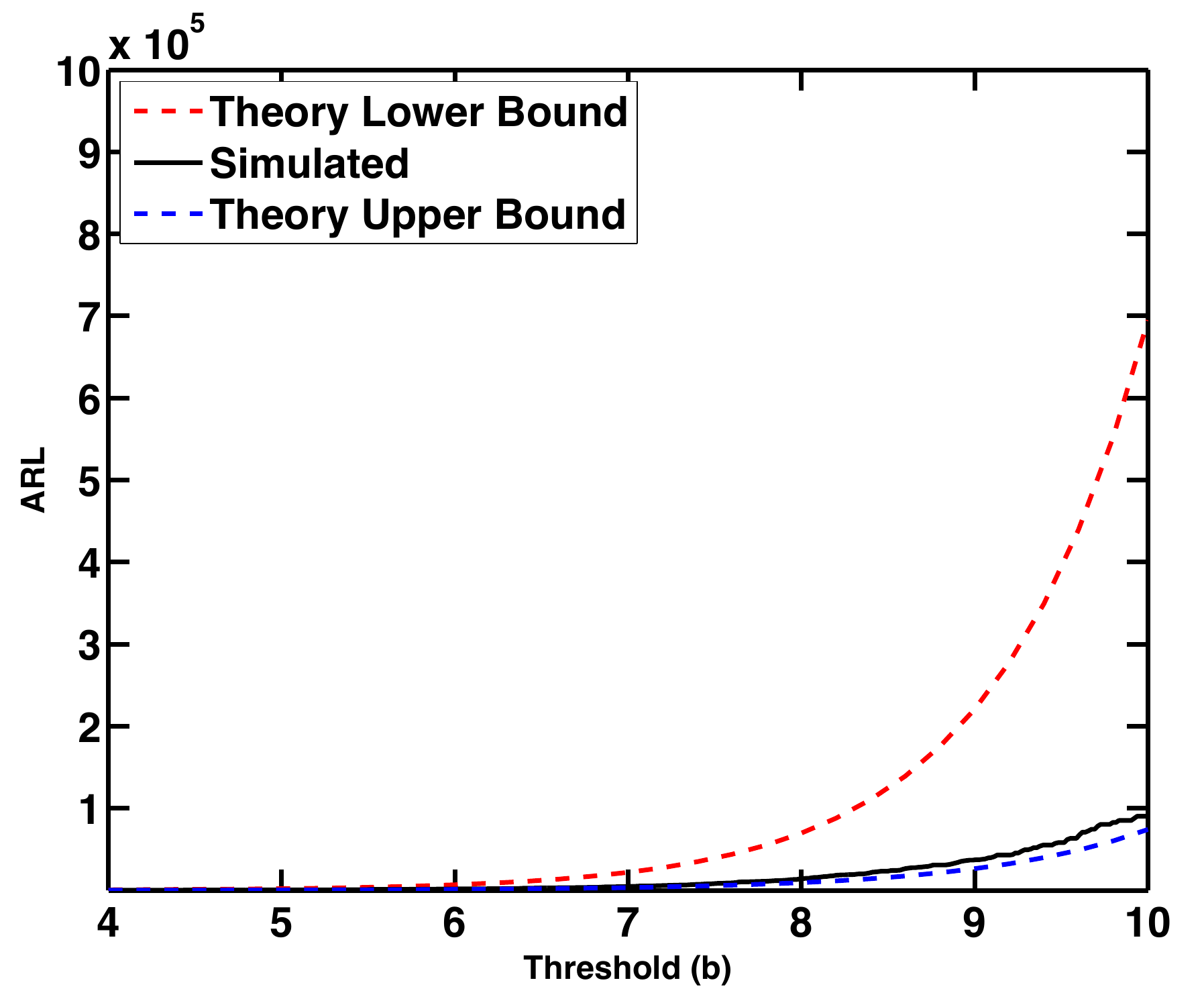} & \includegraphics[width=0.45\linewidth]{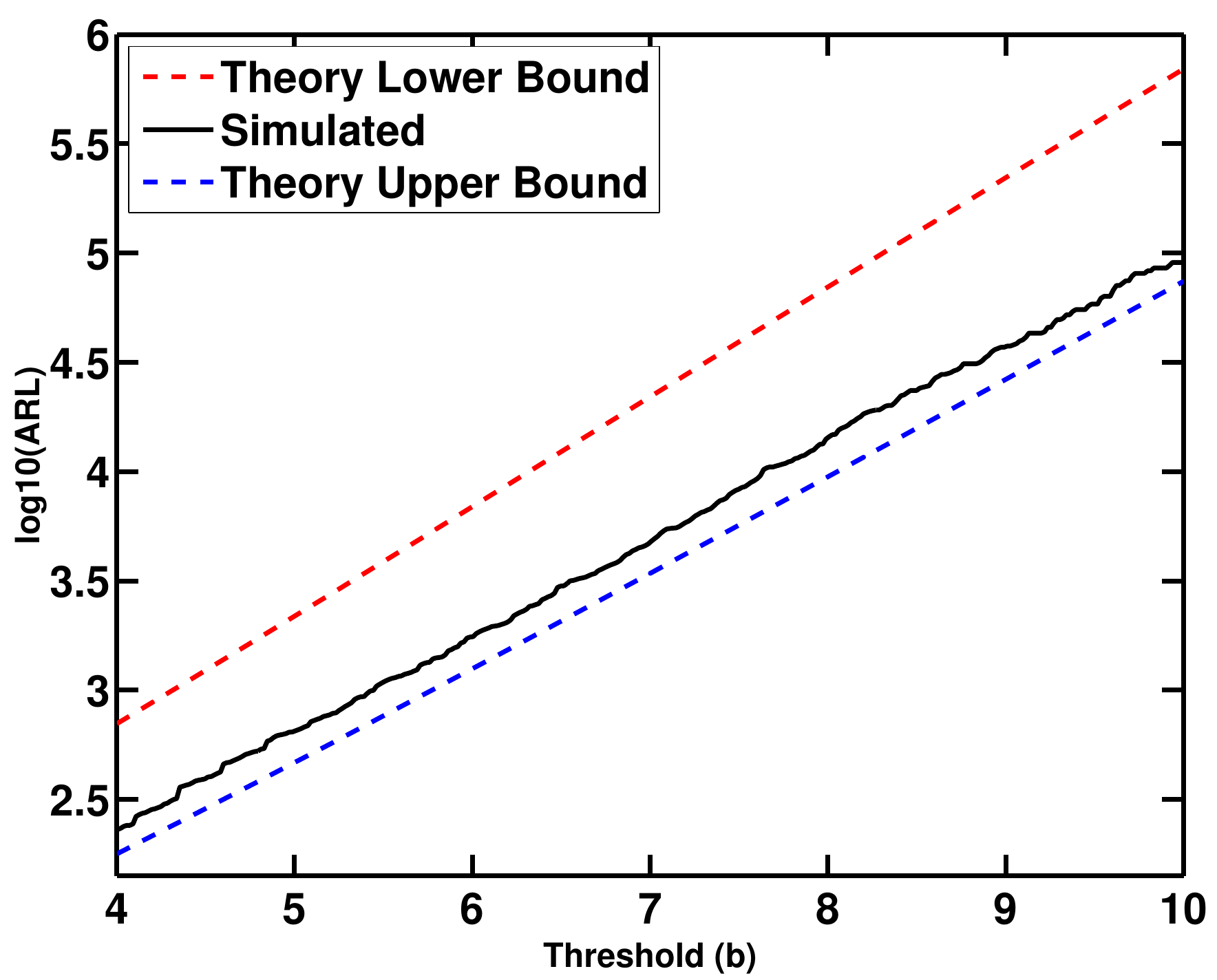} \\
(a) & (b)
\end{tabular}
\end{center}
  \caption{Comparison of the theoretical upper and lower bounds with the simulated ARL for a case with $N=6$, $p_0 = 0.3$, and $p_1 = 0.8$. 
  }
  \label{fig:theory_vs_actual}
\end{figure}
\begin{figure}[h!]
\begin{center}
\begin{tabular}{cc}
\includegraphics[width = 0.45\linewidth]{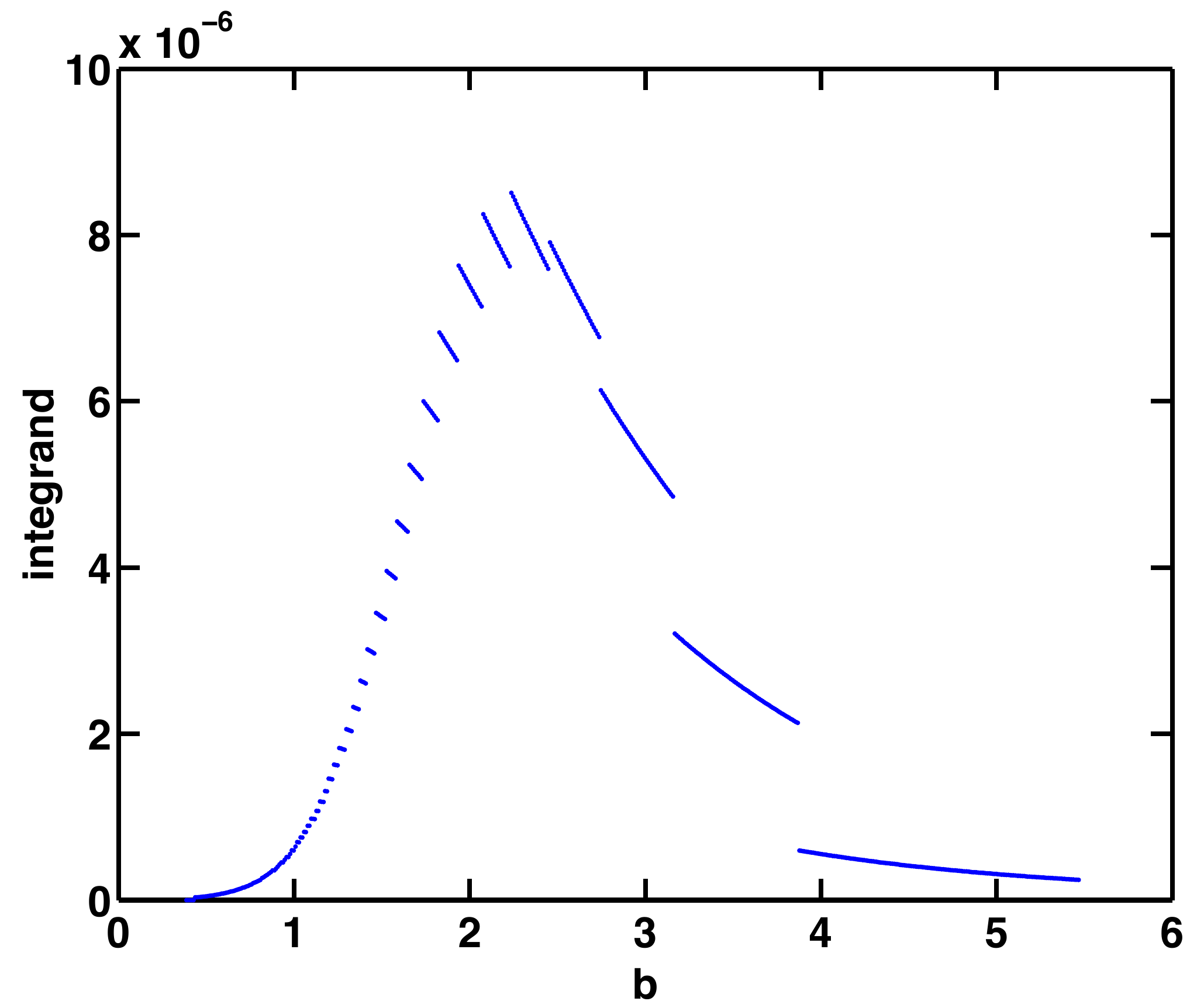} & 
\includegraphics[width = 0.45\linewidth]{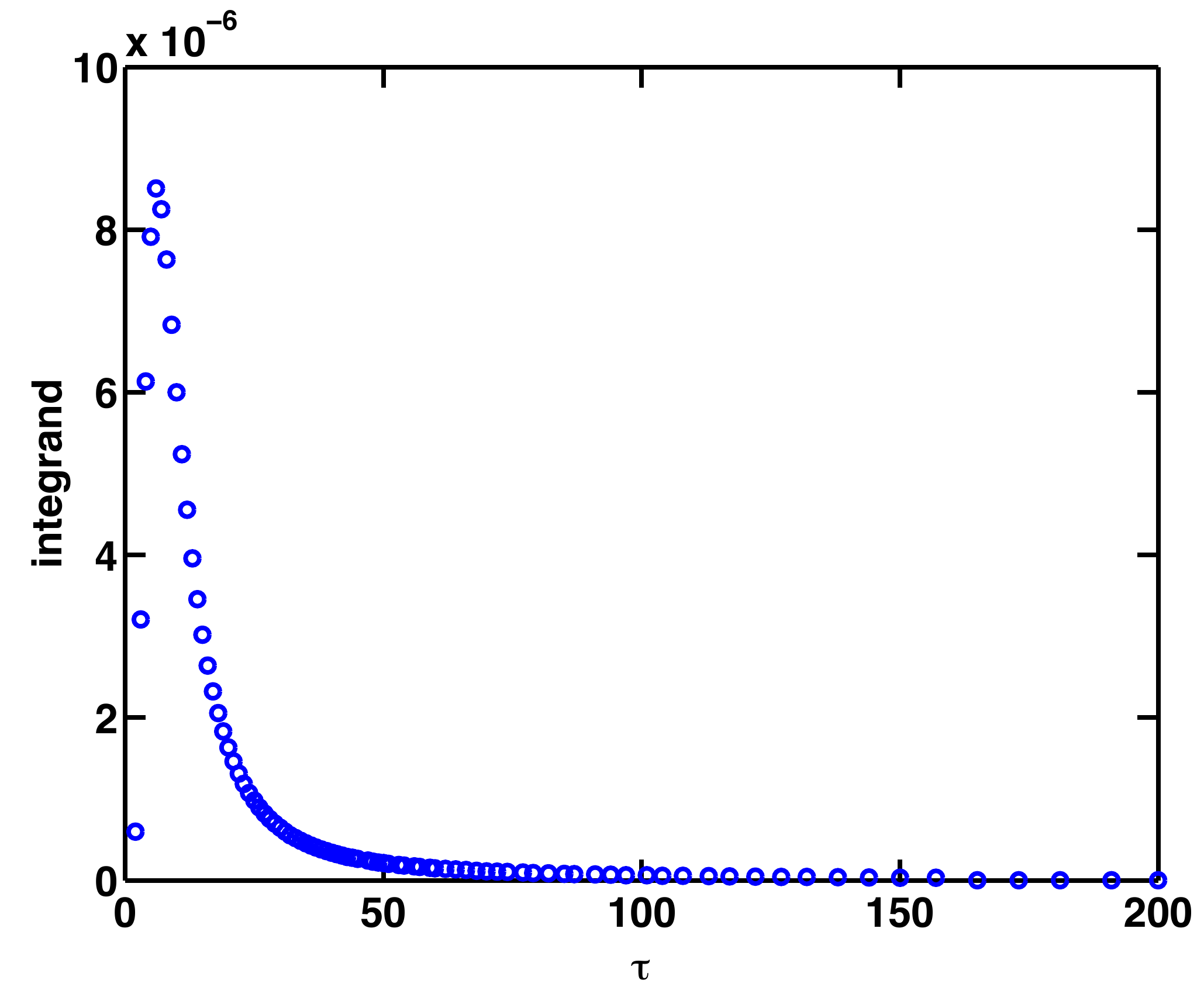} \\ (a) & (b)
\end{tabular}
\end{center}
\caption{For a case $b=7.041$, $p_0=0.3$, $p_1 = 0.8$, and $N=6$, (a): value of the integrand 
in (\ref{eq:ARL_integration}),  and (b): value of the terms in the sum in (\ref{ARL_LB}). Note that only a small subset of these values contribute to the integration or sum in these expressions, and these values correspond to when $\tau$ is relatively small (i.e., hypothesized changepoint location $k$ is not too far away from the current time $t$). 
}
\label{fig:integrand}
\end{figure}

We verify the theoretical upper and lower bounds for ARL of the mixture method versus the simulated values, and consider a case with $p_0 = 0.3$, $p_1 = 0.8$, and $N = 6$. The comparison results are shown in Figure~\ref{fig:theory_vs_actual}, and listed in Table~\ref{table:desired_arl_table}. These comparisons show that the lower bound is especially a good approximation to the simulated ARL and, hence, it can be used to efficiently determine a threshold corresponding a desired ARL (which is typically set to a large number around 5000 or 10000), as shown in Table \ref{table:theo}. 
%
%
Figure \ref{fig:integrand} demonstrates that only small $\tau$ values in the integration equation (\ref{eq:ARL_integration}) contribute to the sum and play a role in determining the ARL. 


\begin{table}[h!]
\begin{center}
\caption{Theoretical vs. Simulated ARL values for $N = 6$, $p_0 = 0.3$, and $p_1 = 0.8$.}
\label{table:desired_arl_table}
\begin{tabular}{|c|c|c|c|}
  \hline
  Threshold & Theory LB   &Theory UB  & Simulated \\ \hline 
  7.3734 & 5000 & 33878 & 6963  \\ \hline
  8.0535 & 10000 & 74309 & 14720  \\ \hline
\end{tabular}
\end{center} 
\end{table}

\begin{table}[h!]
\begin{center}
\caption{Theoretical vs. simulated thresholds for $p_0 = 0.3$, $p_1 = 0.8$, and $N = 6$. The threshold $b$ calculated using theory is very close to the corresponding threshold obtained using simulation.}
\label{table:theo}
\begin{tabular}{|c|c||c|c|}
  \hline
  ARL & Theory $b$  & Simulate ARL & Simulated $b$\\ \hline 
  5000 & 7.37 & 5049 &  7.04 \\ \hline
  10000 & 8.05 & 10210 &  7.64 \\ \hline
\end{tabular}
\end{center} 
\end{table}

\section{Numerical Examples}\label{sec:num_eg}

In this section, we compare the performance of our three methods via numerical simulations. We first use simulations to determine the threshold $b$ for each method, so these methods all have the same average run length (ARL) which is equal to 5000, and then estimate the detection delays using these $b$'s under different scenarios. The results are shown in Table \ref{table:full_search_table}, including the detection delay and the thresholds (shown in brackets). Note that the low-complexity mixture and  H-Mix methods can both detect the community quickly. 


\begin{table}[h!]
\begin{center}
\caption{Comparison of detection delays for various cases of $s$, $N$, $p_0$, and $p_1$. The numbers inside the brackets are the threshold $b$ such that ARL = 5000.}
\label{table:full_search_table}
\begin{tabular}{|m{1.3cm}|m{1.2cm}|m{1.3cm}|m{1cm}|m{1.6cm}|}
  \hline
  Parameters & ES & Mixture ($p_1$known) & H-Mix & Mixture ($p_1$unknown)\\ \hline 
  $p_0 = 0.2$ $p_1 = 0.9$ $s=3$ $N = 6$ &  3.8 (9.96)  & 4.3 (6.71) & 3.8 (9.95) & 9.1 (3.03)\\ \hline
  $p_0 = 0.3$ $p_1 = 0.7$ $s=3$ $N = 6$ & 9.5 (10.17) & 12.8 (6.77) & 10.8 (10.18) & 12.5 (2.94) \\ \hline
  $p_0 = 0.3$ $p_1 = 0.7$ $s=4$ $N = 6$ & 5.0 (8.48)  & 6.7 (6.88) & 6.4 (10.17) & 7.7 (2.03)\\ \hline
\end{tabular}
\end{center}
\end{table}

Next, we test the case when there are a few active edges inside the network that do not form a community, as shown in Fig. \ref{fig:mixture_model_failure_4}(b). Assume the parameters are $p_0 = 0.2$, $p_1 = 0.9$, $s = 3$, and $ N = 6$, and we set the thresholds for each method so they have the same ARL which is equal to 5000. Table \ref{table:mixture_model_failure_table} demonstrates that the ES and H-Mix methods can both identify this ``false community'' case by having relatively long detection delay and, hence, we can effectively rule out such ``false communities'' by setting a small window size $m_1$. In contrast, the mixture method cannot identify a ``false community'' and it  (falsely)  raises an alarm quickly. Code for implementing theoretical calculation and simulations can be found at http://www2.isye.gatech.edu/$\sim$yxie77/CommDetCode.zip.

\begin{table}[h!]
\caption{ARL and detection delay for each algorithm under the conditions $p_0 = 0.2$, $p_1 = 0.9$, $s = 3$, and $N = 6$, where the ARL is 5000 and there are three active edges according to Figure~\ref{fig:mixture_model_failure_4}(b) that do not form the community.}
 \label{table:mixture_model_failure_table}
\begin{center}
\begin{tabular}{|l|l|l|}
  \hline
  Method & Threshold (ARL = 5000) & Detection Delay \\ \hline 
  ES &  9.96 & 49.7  \\ \hline
  Mixture  &  6.71 & 4.3  \\ \hline
  H-Mix  & 9.95 &   100.7   \\ \hline
\end{tabular}
\end{center}
\end{table}

\section{Conclusions and Future Work}\label{sec:con}

In this paper, we have presented and studied three methods for quickly detecting emergence of a community using sequential observations: the exhaustive search (ES), the mixture, and the hierarchical mixture (H-Mix) methods. These methods are derived using sequential changepoint detection methodology based on sequential likelihood ratios, and we use simple Erd\H{o}s-Renyi models for the networks. The ES method has the best performance; however, it is exponentially complex and is not suitable for the quick detection of a community. The mixture method is computationally efficient, and its drawback of not being able to identify the ``false community'' is addressed by our H-Mix method by incorporating a dendrogram decomposition of the network. We derive accurate theoretical approximations for the average run length (ARL) of the mixture method, and also demonstrated the performance of these methods using numerical examples. 
%

Though community detection has been the focus of this paper, locating the community within the network (community localization) can still be accomplished by the exhaustive search and hierarchical mixture methods (find the subgraph with the highest likelihood). 
The mixture method cannot be directly used for localizing the community.
%

We focused on simple Erd\H{o}s-Renyi graphical models in this paper. Future work includes extending our methods to other graph models such as stochastic block models \cite{gomezgardenes06}.


\appendix

\noindent\begin{proof}[Outline of proof for Theorem \ref{thm1}]
We first approximate the distribution of $U_{k, t}^{(i,j)}$ in (\ref{U_def}) by a normal random variable with the same mean and variance. Following (\ref{U_def}),
\[U_{k, t}^{(i,j)} 
=   \log \left( \frac{p_1}{1 - p_1} \frac{1- p_0}{p_0}\right) \left(\sum_{m=k+1}^t [X_m]_{ij}\right) + (t - k) \log \left( \frac{1 - p_1}{1 - p_0} \right) \]
\[=   (c_0 - c_1) \sum_{m=k+1}^t [X_m]_{ij} + (t-k) c_1.\]
%
Let $\tau \triangleq t - k$ denote the number of samples after the hypothesized changepoint location $k$.  Recall that $[X_m]_{ij}$ is a Bernoulli random variable. We will approximate the distribution of $\sum_{m=k+1}^t [X_m]_{ij} $ by a normal random variable with the same mean $p_0 \tau$ and variance $p_0 (1-p_0) \tau$ (when $\tau$ is large, by the central limit theorem this approximation will be accurate).
 Therefore,
\begin{equation}
U_{k, t}^{(i,j)} \approx   \tau[c_1 +  (c_0 - c_1) p_0 ] + 
 Z_{ij} \sqrt{\tau (c_0 -c_1)^2 p_0 (1-p_0)},
\end{equation}
where $Z_{ij}$ are i.i.d. standard normal random variables with zero mean and unit variance. 
Using this normal approximation, the detection statistic $h(U_{k, t}^{(i, j)})$ for the mixture method in (\ref{T_mix}) can be redefined as
\begin{equation}
h_{\tau}(Z_{ij})   \triangleq h (U_{t-\tau, t}^{(i, j)}) 
 = h (g_{\tau} (Z_{ij})).
\end{equation}
%
%
%
Use the equation (3.3)
in \cite{SiegmundYakirZhang2011} for the tail probability of a sum of functions of standard normal random variables, we obtain that
\begin{equation}
\begin{split}
&\mathbb{P}^\infty\{T_{\rm mix}\leq m\}  
 =  \mathbb{P}^\infty \{\max_{t\leq m, m_0\leq t-k\leq m_1}
\sum_{1\leq i < j \leq N}  h_{\tau} (Z) \geq b
\}\\
 \sim & \sum_{\tau = m_0}^{m_1} m \left(1-\frac{\tau}{m}\right) 
\frac{e^{-N[\theta_\tau \dot{\psi}_\tau(\theta_\tau) - \psi_\tau(\theta_\tau)]}}{[2\pi N \ddot{\psi}_{\tau}(\theta_\tau)]^{1/2}} \cdot \theta_\tau^{-1} \gamma^2(\theta_\tau) \left(\frac{N}{\tau}\right)^2 \\
& \cdot
\nu^2\left(\sqrt{\frac{2N\gamma(\theta_\tau)}{\tau}}\right).
\end{split}
\label{eqn20}
\end{equation}
 Note that the above expression (\ref{eqn20}) can be rewritten as
\begin{equation}
\begin{split}
&\sum_{\tau = m_0}^{m_1} 
\frac{m^2}{\tau^2} (1-\frac{\tau}{m}) 
\frac{e^{-N[\theta_\tau \dot{\psi}_\tau(\theta_\tau) - \psi_\tau(\theta_\tau)]}}{[2\pi N \ddot{\psi}_\tau(\theta_\tau)]^{1/2}} \cdot
\theta_\tau^{-1} \gamma^2(\theta_\tau) N^2  \\
&\cdot
\nu^2(\sqrt{\frac{2N\gamma(\theta_\tau)}{\tau}})\cdot \frac{1}{m}.
\end{split}
\label{eqn:sum}
\end{equation}
Let $w = \tau/m$. Hence 
\begin{equation}
\tau = mw. \label{tau_redef}
\end{equation}
When $m$ is large, we can approximate this sum by an integration. We will omit the subscript of $\theta_{\tau}$ and $\psi_\tau$ for simplicity and recover them in the final expression (but it should be kept in mind that they depend on the variable and should be kept inside the integral). Then (\ref{eqn:sum}) becomes
\begin{equation}
N^2 \int_{m_0/m}^{m_1/m} 
\frac{e^{-N[\theta\dot{\psi}(\theta) - \psi(\theta)]}}{[2\pi N \ddot{\psi}(\theta)]^{1/2}}
\theta^{-1} \gamma^2(\theta) \cdot 
 \frac{(1-w)}{w^2} \nu^2(\sqrt{\frac{2 N\gamma(\theta) }{mw}}) dw. \label{sum_2}
\end{equation}
Note that $1-w\approx 1$ since $m$ is much larger than $m_1$ or $m_0$. Using a change-of-variable 
\begin{equation}
y = \sqrt{2N/(mw)}, \label{y_def}
\end{equation}
 we have $w = 2N/(my^2)$ and $dw = (2N)/m\cdot (-2)/(y^3) dy$. Plugging these into the above expression, we have that (\ref{sum_2}) becomes
\begin{equation}
\begin{split}
& N^2 \int_{\sqrt{2N/m_0}}^{\sqrt{2N/m_1}} \frac{e^{-N[\theta\dot{\psi}(\theta) - \psi(\theta)]}}{[2\pi N \ddot{\psi}(\theta)]^{1/2}}
\theta^{-1} \gamma^2(\theta)
\cdot \frac{m^2 y^4}{4N^2} \nu^2(y\sqrt{\gamma(\theta)}) \\
& \cdot 
\frac{2N}{m} \cdot \frac{-2}{y^3} dy \\
& =  m \int_{\sqrt{2N/m_1}}^{\sqrt{2N/m_0}} 
\frac{e^{-N[\theta\dot{\psi}(\theta) - \psi(\theta)]}}{[2\pi  \ddot{\psi}(\theta)]^{1/2}}
\theta^{-1} \gamma^2(\theta) \sqrt{N} y \nu^2(y\sqrt{\gamma(\theta)}) dy
\end{split}
\label{eqn26}
\end{equation}
which is linear in $m$ and a desirable result for our analysis. 

On the other hand, we assume that $T_{\rm mix}$ is exponentially distributed with mean $\lambda$. Hence, $\mathbb{P}^{\infty}\{T\leq m\} = 1-e^{-m/\lambda} \approx m/\lambda$ (from Taylor expansion approximation). Therefore, $\mathbb{E}^\infty[T_{\rm mix}]  = \lambda$. Now combine this with (\ref{eqn26}), we have the ARL is given by 
\begin{equation}
\begin{split}
&\mathbb{E}^\infty[T_{\rm mix}]  \leq \mbox{ARL}_{\rm UB}     \\
= & [\int_{\sqrt{2N/m_1}}^{\sqrt{2N/m_0}} 
\frac{e^{-N[\theta\dot{\psi}(\theta) - \psi(\theta)]}}{[2\pi  \ddot{\psi}(\theta)]^{1/2}} \cdot\theta^{-1} \gamma^2(\theta) \sqrt{N} y \nu^2(y\sqrt{\gamma(\theta)}) dy]^{-1},
\end{split}
\end{equation}
where the last line is equal to (\ref{eq:ARL_integration}).
Note that $\theta_y$ is the solution to $\dot{\psi}_{2N/y^2}(\theta_y) = b/N,$
since $\tau = mw$ and $w = 2N/(my^2)$ by our previous change-of-variable (\ref{tau_redef}) and (\ref{y_def}). This expression can be used as an upper bound to the ARL since it replaces a sum with integral. 

We can also obtain a lower bound to the ARL as follows. Using the mapping $\tau = mw = 2N/y^2$, and equivalently $y = 2N/\tau^2$, we can obtain a lower bound by adapting (\ref{eq:ARL_integration}) into a series, which leads to (\ref{ARL_LB}).
\end{proof}

\bibliography{online_community_detection_bibliography}

\end{document}